\documentclass[sigconf]{acmart} 

\usepackage{algorithm}
\usepackage{algpseudocode} 
\usepackage{setspace} 
\usepackage[normalem]{ulem} 
\usepackage{amsmath}
\DeclareMathOperator*{\argmax}{argmax}

\AtBeginDocument{%
  }

\setcopyright{acmlicensed}
\copyrightyear{2025}
\acmYear{2025}
\acmConference[OARS ’25]{Proceedings of the 5th International Workshop on Online and Adaptive Recommender Systems (OARS ’25), co-located with the 31st ACM SIGKDD Conference on Knowledge Discovery and Data Mining (KDD ’25)}{August 3-7, 2025}{Toronto, Canada}
\acmBooktitle{Companion Proceedings of KDD ’25: Online and Adaptive Recommender Systems (OARS ’25)}

\begin{document}
\title{Contextual Bandits in Payment Processing: Non-uniform Exploration and Supervised Learning}

\author{Akhila Vangara}
\authornote{Both authors contributed equally to this research.}

\affiliation{%
  \institution{Adyen}
  \country{Netherlands}
}

\author{Alex Egg}
\authornotemark[1]
\orcid{0009-0000-8644-9818}
\affiliation{%
  \institution{Adyen}
  \country{Netherlands}
}

\begin{abstract}
Uniform random exploration in contextual bandits supports off-policy learning via supervision but incurs high regret, making it impractical for many applications. Conversely, non-uniform exploration offers better immediate performance but lacks support for off-policy learning. Recent research suggests that regression oracles can bridge this gap by combining non-uniform exploration with supervised learning. In this paper, we analyze these approaches within a real-world industrial context at Adyen, a large global payments processor characterized by batch logged delayed feedback, short-term memory, and dynamic action spaces under the Empirical Risk Minimization (ERM) framework. Our analysis reveals that while regression oracles significantly improve performance, they introduce challenges due to rigid algorithmic assumptions. Specifically, we observe that as a policy improves, subsequent generations may perform worse due to shifts in the reward distribution and increased class imbalance in the training data. This effect arises when regression oracles influence probability estimates and the realizability of subsequent policy models, leading to fluctuations in performance across iterations. Our findings highlight the need for more adaptable algorithms that can leverage the benefits of regression oracles without introducing instability in policy performance over time.
\end{abstract}

\maketitle

\section{Introduction}

Adyen is a global payments processor with a diverse product suite including tools to optimize transaction authorization rates. Various interventions, conditioned on context, can be applied at payment-time to boost transaction authorization rates, naturally framing this problem in a contextual bandit setting where interventions correspond to actions and authorization feedback from the bank (environment) serve as rewards. It is instructive to view this system in the framework of a recommender system.

In many industrial recommender system applications, learning from historical logs---often collected under biased feedback conditions---is crucial. This paper investigates how traditional supervised learning, via Empirical Risk Minimization (ERM), can be combined with decision-making in interactive settings as formalized by contextual bandits. Our aim is to leverage supervised signals from logged data to inform more effective exploration strategies.

Empirical Risk Minimization (ERM) underpins traditional supervised learning. In the ERM framework, a learning algorithm is given a dataset
\[
\mathcal{D} = \{(x_i, y_i)\}_{i=1}^n,
\]
where each instance \(x_i\) is paired with a fully observed ground-truth label \(y_i\). The goal is to learn a hypothesis \(h \in \mathcal{H}\) that minimizes the empirical risk,
\[
\hat{R}(h) = \frac{1}{n} \sum_{i=1}^{n} \ell(y_i, h(x_i)),
\]
with \(\ell: \mathcal{Y} \times \mathcal{Y} \to \mathbb{R}_+\) quantifying the discrepancy between the predicted and true labels. This framework assumes that the training data is both complete and unbiased---a \emph{full-information} setting. In contrast, many real-world systems rely on data collected through biased logging policies, which may cause ERM to inadvertently learn these biases instead of the true underlying relationships.

Contextual bandits introduce a setting where, at each round, an algorithm observes a context \(x \in \mathcal{X}\) and must choose an action \(a \in \mathcal{A}\) from a (finite or continuous) action set. Once an action is taken, the algorithm receives a reward \(r \in \mathbb{R}\), but it observes feedback \textit{only for the chosen action}. Formally, at each round \(t\) the learning process is as follows:
\begin{enumerate}
    \item Receive a context \(x_t \in \mathcal{X}\).
    \item Choose an action \(a_t \in \mathcal{A}\) according to a policy \(h: \mathcal{X} \to \mathcal{A}\).
    \item Observe the reward \(r_t = r(x_t, a_t)\) associated with the chosen action.
\end{enumerate}
The goal is to develop a policy that maximizes cumulative reward (or equivalently minimizes regret relative to the best policy). Two challenges arise from this formulation:
\begin{itemize}
    \item \textbf{Exploration vs. Exploitation:} Since the algorithm sees only the reward of the chosen action, it must balance exploiting actions with high estimated rewards and exploring less-certain actions to acquire additional information.
    \item \textbf{Partial Feedback:} Unlike full-information settings, the learner receives only partial feedback, making standard ERM techniques directly inapplicable.
\end{itemize}

A common exploration strategy in this setting is the \(\epsilon\)-greedy policy. Under this approach, the learner selects the action with the highest estimated reward with probability \(1-\epsilon\), and chooses an action \textit{uniformly at random} with probability \(\epsilon\). Formally, for a given context \(x\) and action set \(\mathcal{A}\), the \(\epsilon\)-greedy policy \(h_\epsilon\) is defined as
\[
h_{\epsilon}(x) = 
\begin{cases}
\displaystyle \arg\max_{a \in \mathcal{A}} R(x, a), & \text{with probability } 1-\epsilon, \\
\text{a random action from } \mathcal{A}, & \text{with probability } \epsilon,
\end{cases}
\]
where \(R(x, a)\) estimates the expected reward for taking action \(a\) in context \(x\).

The \(\epsilon\)-greedy exploration strategy has no specific modeling assumptions which means it can easily leverage existing ML infrastructure: a la regression models (e.g., gradient boosting or NNs). This makes it very popular in industry \cite{grubhub,Zhu2009RevenueOW,adyen-badits}), however, its linear regret due to the reliance on \textit{uniform random exploration} is an expensive trade off. 

Of course, contextual bandits literature has made progress against the regret issue for years and many solutions have been proposed however, they all come with limitations. For instance, early methods like Upper Confidence Bound (UCB/1) and Thompson Sampling enjoyed non-uniform exploration, but either lack adequate contextual integration or require heavy computational overhead and rigid modeling assumptions. LinUCB~\cite{LinUCB} improved upon this by incorporating linear models with analytical uncertainty bounds, though the linear model introduces context,  its linearity limits its expressiveness (modeling assumptions). Subsequent approaches, such as NeuralUCB~\cite{NeuralUCB} and NNLinUCB~\cite{NNLinUCB}, relax the linearity assumption but incur high computational costs or perform only “shadow” exploration on neural network features. In contrast, methods like EE-Net~\cite{EE-Net} employ multiple neural networks to decouple exploitation from exploration, achieving better performance in some settings but still have rigid modeling assumptions.

A recent line of work further advances exploration by introducing \textit{regression oracles}~\cite{foster18a,SquareCB}. These methods transform supervised learning predictions into online policies, thereby harnessing the power of general-purpose supervised algorithms (e.g., neural networks or boosting) to address the exploration--exploitation trade-off.

In this work, we build on these ideas by proposing a methodology that achieves non-uniform exploration in contextual bandits with standard supervised models.

\textit{\textbf{Our contributions are novel applications of regression oracles that enable}}:
\begin{itemize}
    \item Non-random uniform exploration without modeling assumptions
    \item Context between arms
    \item   Logged bandit feedback
\end{itemize}

\begin{table}[h]
\centering
\begin{tabular}{|l|c|c|c|}
\hline
\textbf{Algorithm} & \textbf{Contextual} & \textbf{Non-uniform} & \textbf{Assumptions*} \\
\hline
UCB1            & \(\times\)             & \(\checkmark\)          & \(\times\) \\
TS              & \(\times\)             & \(\checkmark\)          & \(\times\) \\
LinUCB          & \(\checkmark\)         & \(\checkmark\)          & \(\times\) \\
NNLinUCB          & \(\checkmark\)         & \(\checkmark\)          & \(\times\) \\
NeuralUCB          & \(\checkmark\)         & \(\checkmark\)          & \(\times\) \\
Epsilon-greedy  & \(\checkmark\)         & \(\times\)              & \(\checkmark\) \\
SquareCB        & \(\checkmark\)         & \(\checkmark\)          & \(\checkmark\) \\
\hline
\end{tabular}
\caption{Comparison of contextual bandit algorithms across key dimensions. *Modeling Assumptions. One can observe the tradeoff between low cost exploration and model assumptions wrt context}
\label{tab:alg_comparison}
\end{table}

\section{Methodology}
Given the introduction to the setting and constraints at Adyen we hope to clearly define our problem space and layout our approach to a solution.

\subsection{Problem Definition}
 The objective is to maximize cumulative reward over time by learning an optimal policy \( \pi \) that maps contexts to actions.

Currently, $\pi$ is $\epsilon$-greedy, using a boosting model during exploitation. However, uniform random exploration in $\epsilon$-greedy yields linear regret; reducing this regret remains an important open problem with substantial practical value.

Motivated from the introduction and the problem, we have 3 requirements:
\begin{itemize}
    \item  Usage of logged bandit feedback
    \item Contextual with no modeling assumptions
    \item Non-uniform random exploration
\end{itemize}

How can we perform non-uniform exploration while leveraging logged bandit feedback? Regression oracles provide a promising direction by connecting supervised learning techniques with contextual bandit algorithms.

\subsection{Regression Oracles}

Regression oracles are black box functions that make real-valued predictions for rewards $\hat{r}_t \in R$  based on context-action pairs $(x_t, a_t)$. After each prediction, the oracle receives the actual reward and updates its internal model $r_t$.
\[
f(x, a) = \mathbb{E}[r \mid x, a],
\]
the expected reward for action \( a \) in context \( x \). The regression oracle selects actions using the induced policy:
\[
\pi_f(x) = \arg\max_{a \in \mathcal{A}} f(x, a),
\]
where \( f \in \mathcal{F} \) is the function learned by the regression oracle to approximate the true reward function. 

\subsubsection{SquareCB}
\begin{figure}
    \centering
    \includegraphics[width=1.0\linewidth]{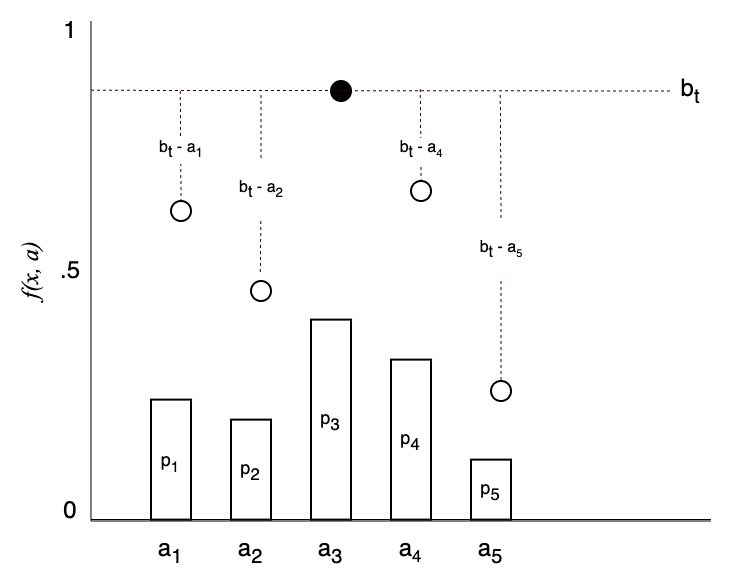}
    \caption{SquareCB Regression Oracle. For \(a  \in \mathcal{A}\), let \(r_a = r(x, a)\) (e.g., using XGBoost for prediction), then build an exploration probability distribution \(p_t\) based on differences \(b_t\).}
    \label{fig:sqcb-diagram}
\end{figure}

Some regression oracles, like SquareCB \cite{SquareCB} are able to explore \textit{non-randomly} which is the key for industrial applications. At each step, it computes predictions for each action, selects the best, assigns probabilities inversely proportional to the gap from the best, \textit{samples} an action, and updates the oracle. This routine is outlined in Algorithm \ref{alg:main}, inspired by Krishnamurthy \cite{SquareCB-lecture}.  The key is the probability selection scheme from Abe \& Long \cite{abe1999associative}, is visualized in Figure \ref{fig:sqcb-diagram}. The key advantage is that it can leverage supervised learning models, such as boosting, for the oracle, which makes it practical for industrial applications.

\begin{algorithm}[h]
  \setstretch{1.1}
  \begin{algorithmic}[1]
    \State \textbf{Parameters:}
    \Statex{}~~~~Learning rate \(\gamma>0\), exploration parameter \(\mu>0\).
    \Statex{}~~~~Offline regression oracle \(SqAlg\) trained over a window of \(L\) days.
    \For{\(t=1,\ldots,T\)}
      \State Receive context \(x_t\).
      \State For each action \(a\in A\), compute
      \[
      \hat{r}_{t,a} = \hat{r}_{t}(x_t,a).
      \]
      \State Let \(b_t = \argmax_{a \in A} \hat{r}_{t,a}\).
      \State For each \(a\neq b_t\), define 
      \[
      p_{t,a} = \frac{1}{\mu + \gamma\left(\hat{r}_{t,b_t} - \hat{r}_{t,a}\right)},
      \]
      and set 
      \[
      p_{t,b_t} = 1-\sum_{a\neq b_t} p_{t,a}.
      \]
      \label{line:prob}
      \State Sample \(a_{t}\sim p_t\) and observe reward \(r_t\sim R(x_t,a_t)\).
    \EndFor
    \State Collect \(T = L\) samples of \(((x_t,a_t),r_t)\) and retrain \(SqAlg\).
  \end{algorithmic}
  \caption{SquareCB: A Regression Oracle-Based Non-Uniform Exploration Algorithm. The algorithm computes action probabilities based on the gap between the best predicted reward \(\hat{r}_{t,b_t}\) and each alternative action \(\hat{r}_{t,a}\), samples an action accordingly, and periodically retrains the offline regression oracle using a batch of logged data.}
  \label{alg:main}
\end{algorithm}

\subsection{Assumptions}

The implementation of regression oracles in large-scale industrial settings, such as at Adyen, introduces unique considerations and challenges. Below, we outline the key assumptions and how they adapt to our setting.

\subsubsection{Delayed Feedback (Offline Oracles)}
A key assumption underlying regression oracles is the availability of \textit{online oracles} that can update incrementally online. However, in many industrial environments---including Adyen---maintaining a system that supports online learning is impractical due to modeling or system constraints (eg updating a tree model incrementally \cite{grubhub}).

Instead, we employ a batch-offline system, where model updates occur in discrete batches rather than in real time. This approach strikes a necessary balance between performance and infrastructural efficiency. Nevertheless, it represents a departure from the theoretical underpinnings, as the model's parameters are updated in a delayed fashion rather than online. This could be viewed in paradigm of \textit{delayed feedback} from reinforcement learning literature.

\subsubsection{Regression}

The SquareCB framework assumes that the functions in the value function class \( F \) are optimized using square-loss regression -- as implied by the name. However, given that our rewards are binary, we employ a binary cross-entropy loss function for our regression oracle. This adaptation aligns the loss function with the binary nature of our reward signals while preserving the oracle's ability to approximate expected rewards.

\subsubsection{Realizability}

Regression oracles operate under a strong realizability assumption, which requires the value function \( r \) chosen by the oracle to closely approximate the true reward distribution. Specifically, for each trial (or in our case, transaction) \( t \), there exists a function \( r^{\ast} \in R \) such that:

\[
r^{\ast}(x, a) = \mathbb{E}[l_{t}(a) \mid x_{t} = x],
\]

where \( r^{\ast} \) represents the true underlying reward function. The oracle's goal is to select a function \( r \) that best approximates \( r^{\ast} \). In practice, however, this assumption is often partially violated because the true reward distribution is complex and hidden. As such, even the best available value function \( r \) may only approximate the true mean reward, introducing potential gaps between theoretical assumptions and practical implementations.

\subsubsection{Dynamic Action Space}

Traditional contextual bandit frameworks assume a static action space \( \mathcal{A} \), where the set of available actions remains fixed. While this is true in our setting, we introduce a \textit{dynamic action space}, where action space (eligible actions) can vary across payment contexts. This dynamic nature increases the complexity for both the \textit{oracle} and the \textit{policy}, as they must account for additional variability when learning and optimizing over the action space.

For each context \(x_t\), the set of available actions \(\mathcal{A}_t\) is determined dynamically using a two-step process:

\begin{enumerate}
    \item \textbf{Rule-based Filtering:} Remove actions that conflict with predefined constraints, such as:
    \begin{itemize}
        \item Merchant payment service provider contracts,
        \item Card network regulations (e.g., country restrictions imposed by Visa/MasterCard),
        \item Transaction amount thresholds.
    \end{itemize}
    
    \item \textbf{Risk-based Pruning:} Exclude actions considered high-risk based on:
    \begin{itemize}
        \item Historical fraud rates for specific merchant, country, and currency combinations,
        \item Real-time monitoring of authorization rates.
    \end{itemize}
\end{enumerate}

This process typically produced $|\mathcal{A}_t| \in [2, 15]$ actions per transaction, with median 4 actions.

\section{Experiments}

In this section, we empirically evaluate our proposed regression oracle approach—implemented via SquareCB—in the context of payment processing at Adyen. We compare its performance against a standard \(\epsilon\)-greedy baseline under realistic industrial settings. Our experiments focus on key metrics such as overall performance, effective exploration, and the impact on subsequent model training.

\subsection{Experimental Setup}

Our evaluation framework consists of three main components:

\begin{enumerate}
    \item \textbf{Data Collection:} We utilize a logged bandit feedback dataset extracted from Adyen’s production system. The dataset consists of 180 million samples collected over a 30-day period. Each sample comprises 56 contextual features (e.g., transaction amount, country, device type), an action label corresponding to the applied intervention, and a binary reward indicating whether the payment was authorized. Table~\ref{tab:dataset_description} provides an overview of the dataset attributes.
    
    \item \textbf{Offline Oracle Training:} We train a regression oracle using the Empirical Risk Minimization (ERM) framework. In our implementation, we employ a boosting classifier with binary cross-entropy loss to approximate the expected reward for each context-action pair (last line of Algorithm \ref{alg:main}.)
    
    \item \textbf{Policy Deployment and A/B Testing:} The trained oracle is integrated into the SquareCB policy to generate non-uniform exploration probabilities (see Algorithm~\ref{alg:main}). For comparison, we deploy a traditional \(\epsilon\)-greedy policy at varying exploration rates (1\%, 4\%, and 6\%) as baselines. Both policies are A/B tested by routing 5\% of the live traffic to each variant over a four-week period.
\end{enumerate}

\begin{table}[h]
    \centering
    \begin{tabular}{|l|l|}
        \hline
        \textbf{Attribute} & \textbf{Description} \\ \hline
        Samples & 180M \\ \hline
        Features & 56\\ \hline
        Target & Authorized (binary) \\ \hline
        Imbalance & 90\% positive \\ \hline
    \end{tabular}
    \caption{Dataset Description for Offline Oracle Training}
    \label{tab:dataset_description}
\end{table}

\subsection{Evaluation Metrics}

We evaluate performance based on the following metrics:
\begin{itemize}
    \item \textbf{Cumulative Reward/Uplift:} The primary metric is the cumulative number of authorized transactions. We report percentage uplifts relative to the \(\epsilon\)-greedy baseline.
    \item \textbf{Effective Exploration Rate:} Defined as the proportion of rounds in which a non-greedy (i.e., non-optimal) action is chosen. This metric helps us assess whether the non-uniform exploration of SquareCB improves the diversity of training data compared to uniform random exploration.
    \item \textbf{Action Diversity:} We quantify diversity using Lorenz curves and the traffic-weighted Gini coefficient, thereby measuring how evenly different interventions are selected across varying context groups.
\end{itemize}

The experiments involved A/B testing models with varying learning rates against $\epsilon$-greedy baselines. The baseline uniform random $\epsilon$-greedy policies were set at three levels of exploration: 1\%, 4\%, and 6\%. Each variant received 5\% of the total traffic to ensure a fair comparison. We conducted experiments over a four-week period.

Point estimates of success probabilities were calculated, along with 75\% and 95\% confidence intervals to measure the robustness of each policy.

\begin{table}[h]
    \centering
    \begin{tabular}{|l|c|c|c|}
        \hline
        \textbf{Experiment} & \textbf{Learning Rate} & \textbf{\% Total Traffic} & \textbf{Duration} \\
        \hline
        Oracle 1       & 1\%  & 5\% & 4 weeks \\
        Oracle 2       & 4\%  & 5\% & 1 week  \\
        Oracle 3       & 6\%  & 5\% & 1 week  \\
        \hline
        \(\epsilon\)-greedy & N/A  & 5\% & 4 weeks \\
        \hline
    \end{tabular}
    \caption{A/B tests with different learning rates and traffic splits.}
    \label{tab:ab_tests_learning_rates}
\end{table}

As noted in the Introduction, we did not benchmark against UCB-style or Thompson-Sampling algorithms because they lack support for both contextual information and non-uniform exploration—two requirements central to our setting. NeuralUCB was likewise omitted due to its substantial computational overhead and additional neural-network modeling assumptions. A summary of these differences appears in Table \ref{tab:alg_comparison}.

Payments come into Adyen - if they pass various risk checks they will be sent to our optimization system where 1 or many interventions will be applied (based on the oracle) and sent to the environment for the reward.

\section{Results}

First we'll look at overall performance and then break it down by exploration and exploitation segments.

\subsection{Overall Performance}

\begin{figure}[h!]
    \centering
    \includegraphics[width=1.0\linewidth]{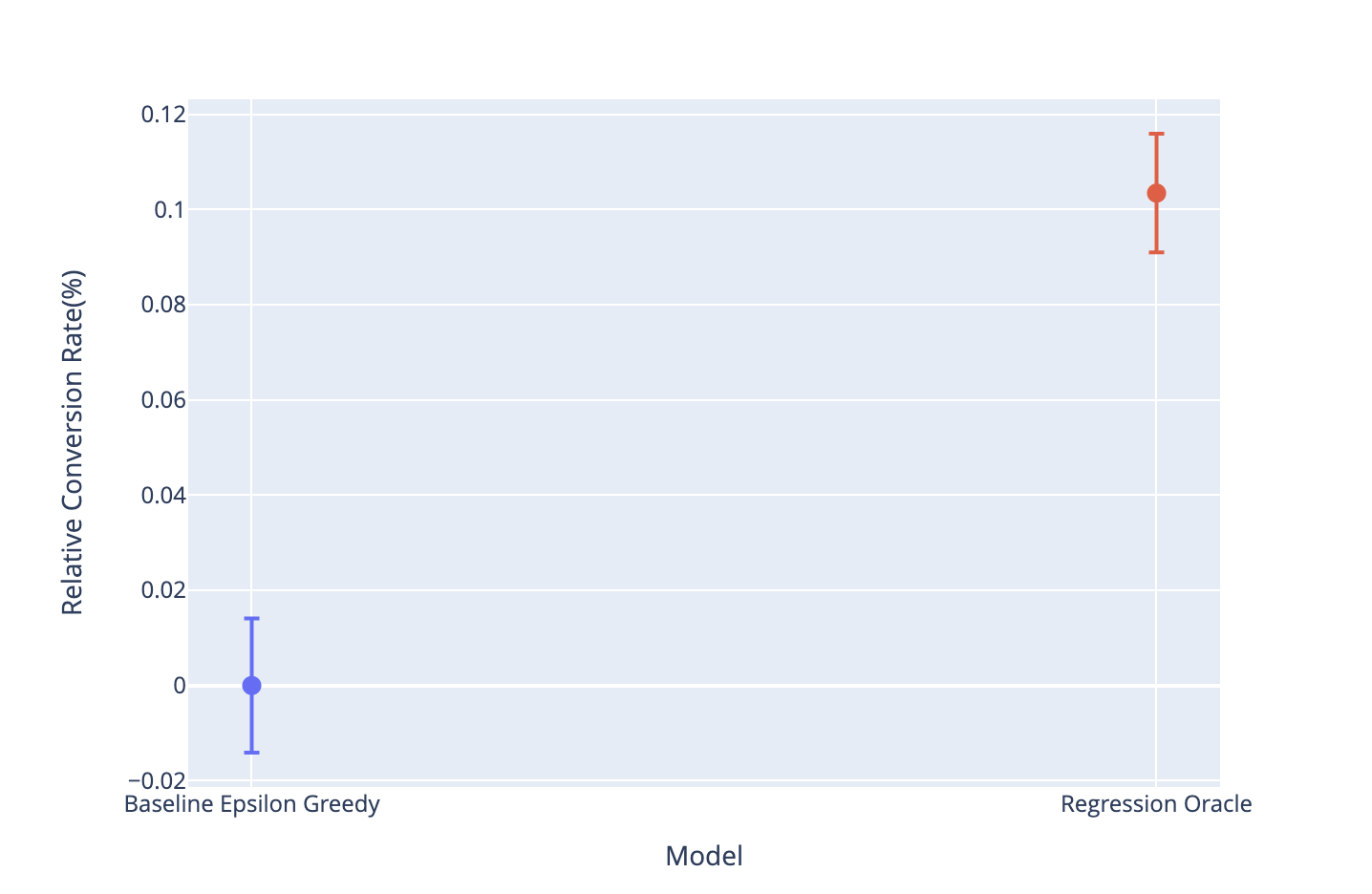}
    \caption{95\% confidence intervals comparing the performance of SquareCB and \(\epsilon\)-greedy policies.}
    \label{fig:performance_baseline}
\end{figure}

Figure~\ref{fig:performance_baseline} shows that the best-performing SquareCB variant achieved a +0.1\% uplift over the baseline \(\epsilon\)-greedy policy within the four-week test period. This improvement translates to an estimated 9 million incremental authorized transactions per year. The confidence intervals (95\%) indicate that the improvement is statistically significant. The intuition behind the performance gains of regression oracles is attributed to the \textit{reduced regret from non-uniform random exploration} -- the premise of this whole line of research.

To confirm our intuition, we measured our effective/actual exploration rates for $\epsilon$-greedy and SquareCB policies respectively in Table \ref{tab:effective-exploration} with the hypothesis that regression oracles would explore \textit{less} but \textit{more efficiently}. However, we observed surprising results: The 1\% $
\epsilon$-greedy policy was exploring only 0.7\% of the time compared to 1.2\% for even the most exploitative SquareCB 50K variant.  These counterintuitive results are explored more in discussion section \ref{sec:eff-exp} . 

\begin{figure}[h!]
    \centering
    \includegraphics[width=1.0\linewidth]{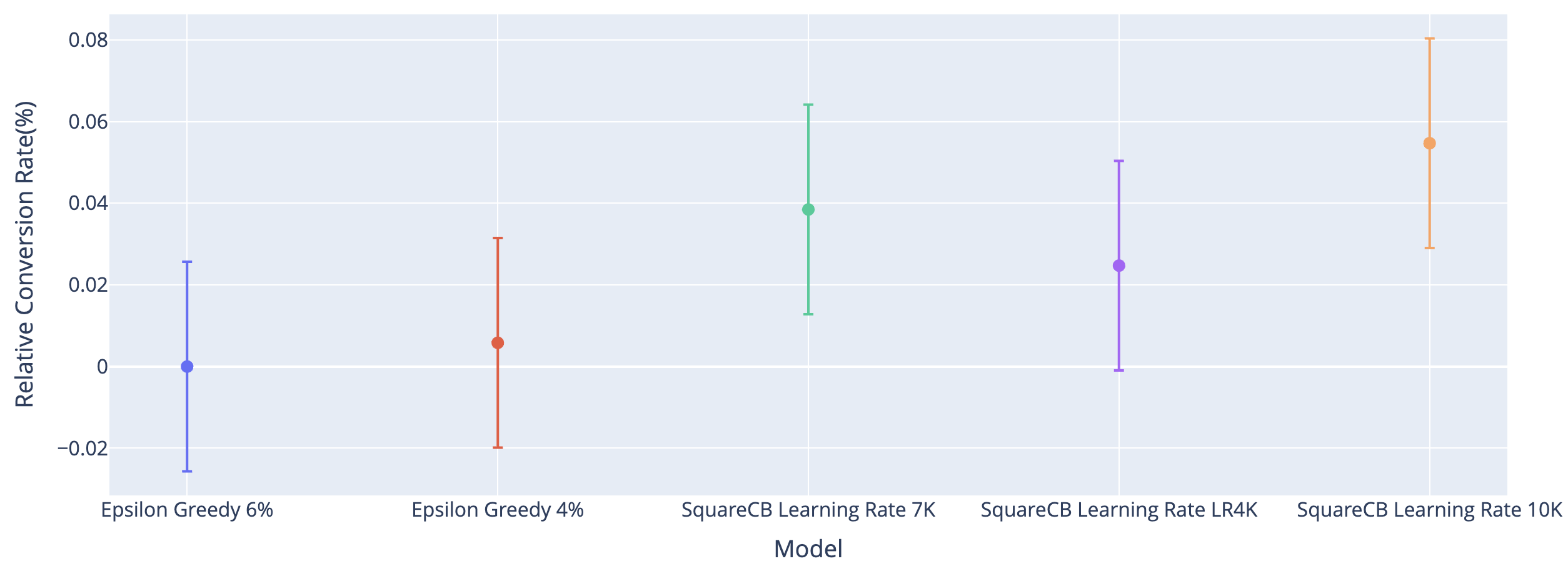}
    \caption{Comparison of $\epsilon$-greedy and SquareCB variants across various exploration rates. Effective exploration rates are in table \ref{tab:effective-exploration}. e-greedy 6\% is 3.45\% and SquareCB 10k is 3.6\% which are essentially the same rate of exploration and as you can see SquareCB variant outperforms. It should be highlighted that for equal exploration rates (3.5\%) the regression oracle variant suffers much less regret.}
    \label{fig:performance_variants}
\end{figure}

Given these results, we wanted to compare $\epsilon$-greedy and SquareCB while controlling for exploration rates. So, in the spirit of fairness, Figure~\ref{fig:performance_variants} illustrates the performance of SquareCB across different learning rates against the $\epsilon$-greedy baselines.

Interestingly enough, controlling for exploration rate, the SquareCB policy (square 10k) always outperforms the baseline (e-greedy 6\%)

\subsection{Exploration vs. Exploitation Trade-off}

Given that we wish to see the benefits of non-random exploration policy, we compare performance on both parts of the traffic: exploration and exploitation. The hypothesis is that we should see marked uplift for the exploration traffic.

\begin{figure}[h!]
    \centering
    \includegraphics[width=1.0\linewidth]{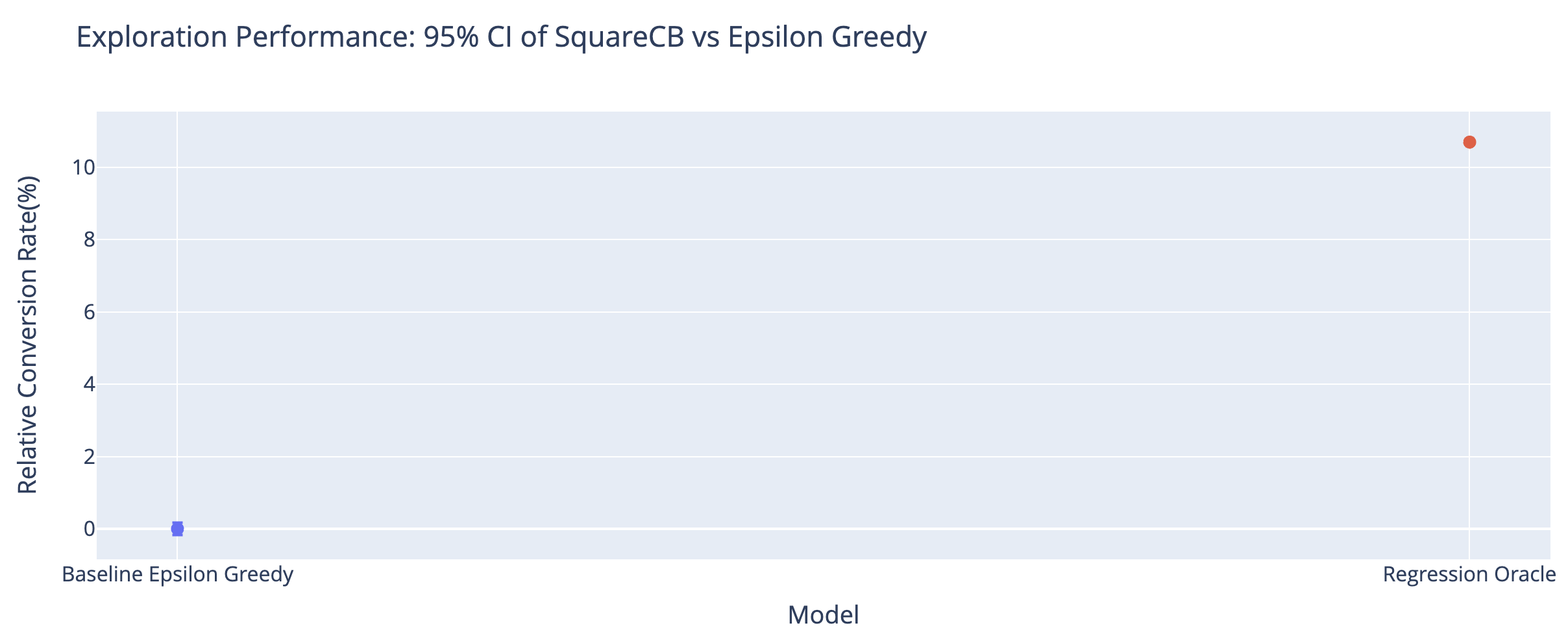}
    \caption{Exploration performance (non-optimal action rounds) comparing SquareCB and \(\epsilon\)-greedy policies.}
    \label{fig:exploration_performance}
\end{figure}

Focusing on the exploration traffic, when the non-optimal action was taken, the SquareCB policy shows a significant \textbf{+11\%} improvement in performance, as it selects actions from a \textit{non-uniform }random distribution via the SqAlg in Algorithm \ref{alg:main}.  Figure~\ref{fig:exploration_performance}.

\begin{figure}[h!]
    \centering
    \includegraphics[width=1.0\linewidth]{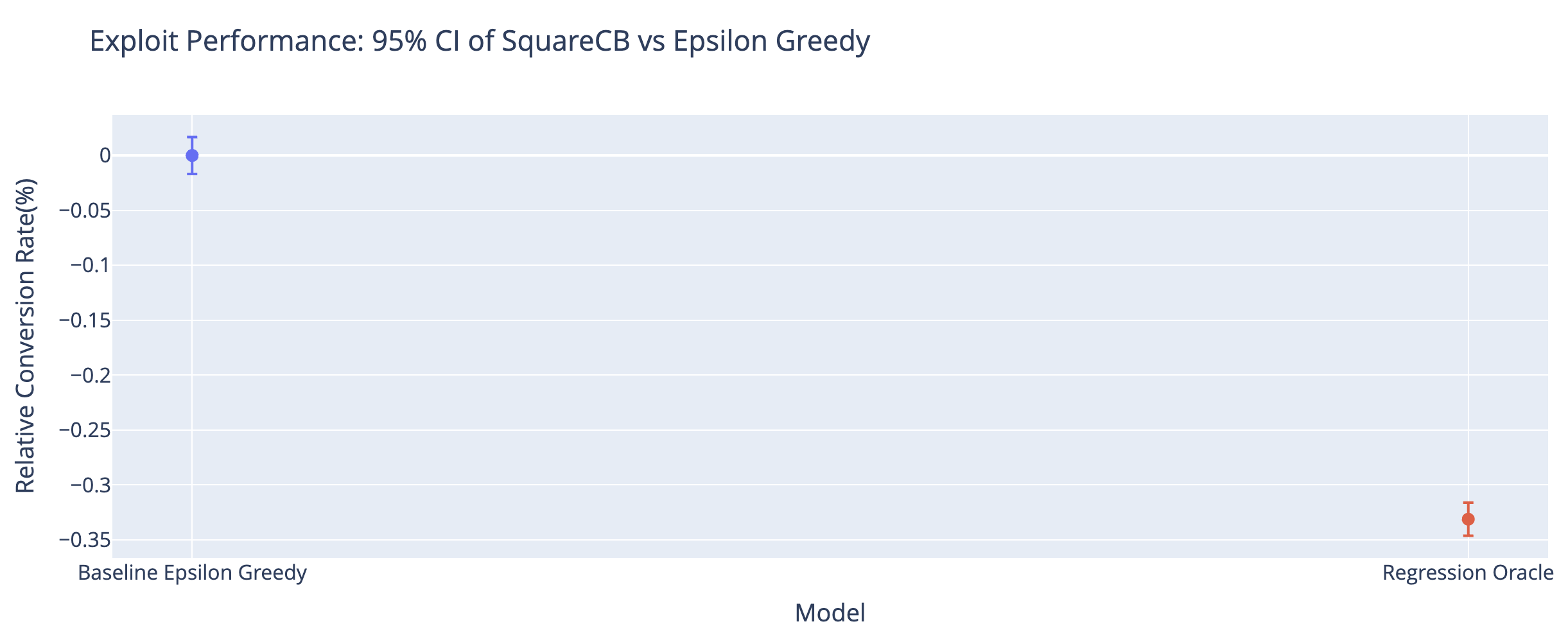}
    \caption{Exploitation performance (optimal action rounds) for SquareCB and \(\epsilon\)-greedy policies.}
    \label{fig:exploitation_performance}
\end{figure}

Now, looking just at the \textit{exploitation} traffic, the results are counter-intuitive as we observed a slight decrease in exploitation performance, with a reduction of \textbf{0.33\%} compared to the baseline policy (Figure~\ref{fig:exploitation_performance}). 

This trade-off is due to two main factors:

\begin{enumerate}
    \item \textbf{Partial Realizability}: The supervised classification algorithm imperfectly models the reward distribution, partially violating the realizability criteria. This highlights the importance of using high-quality classifiers, as a suboptimal classifier can increase regret, as noted in the original theoretical framework.
    \item \textbf{Action Probability Distribution}: Exploration and exploitation are no longer entirely random. Traffic where multiple actions have closely clustered probabilities tends to be explored more often, while traffic with a single clear "best" action (with a high probability gap between it and the others) is exploited. In our setup, this often corresponds to scenarios with larger action spaces ($|A| > 2$), where performance expectations are naturally lower due to increased variability.
\end{enumerate}

Despite the slight loss in exploitation, the gains in exploration far outweigh this trade-off, \textit{resulting in an overall improvement in policy performance} (Figure \ref{fig:performance_baseline}).

In addition to the strong performance we found some interesting insights as we analyzed the experimental results more closely.

\subsection{Exploration Across Dynamic Action Spaces}
\label{sec:eff-exp}

\begin{figure}[h!]
    \centering
    \includegraphics[width=1.0\linewidth]{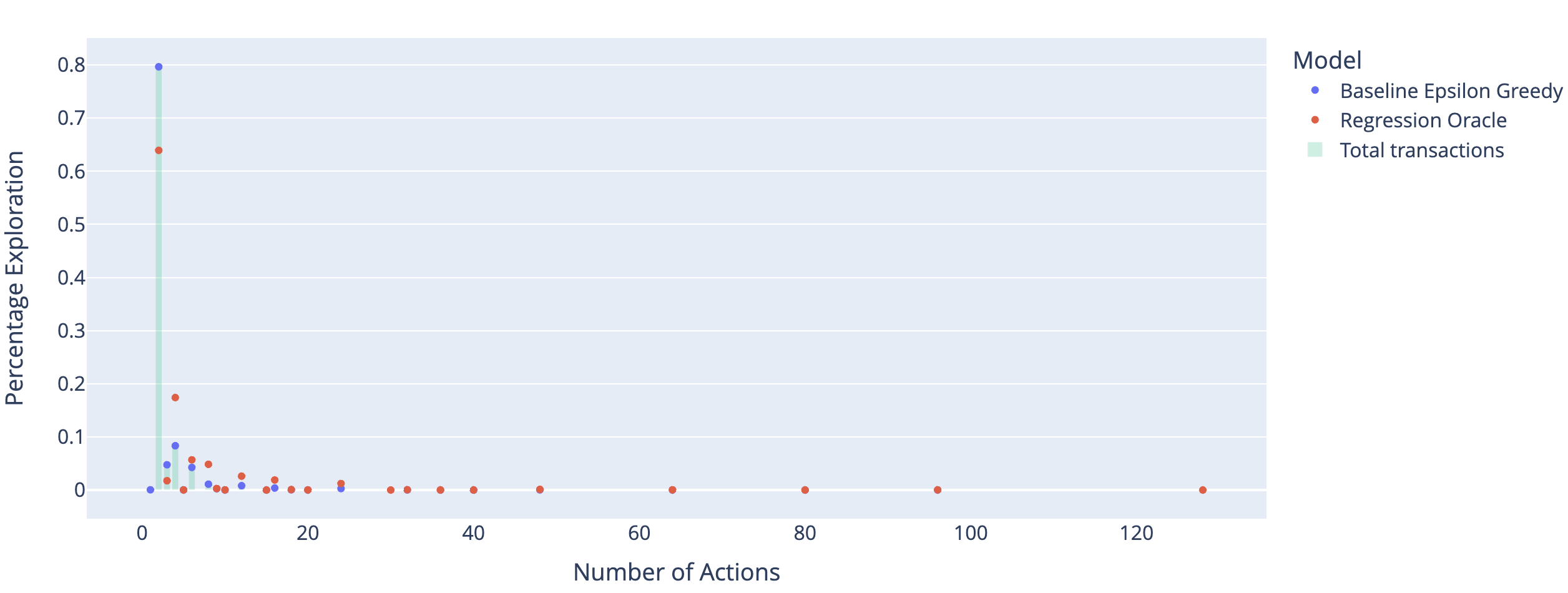}
    \caption{Exploration rates across action space sizes for $\epsilon$-greedy and SquareCB policies. One can observe that for action space sizes greater than 3 the SquareCB policy starts to explore more frequently.}
    \label{fig:exploration-action-size}
\end{figure}

In dynamic action spaces, adequate exploration across varying action space sizes is challenging and nuanced.  $\epsilon$-greedy policies are invariant to action space and explore uniformly regardless of size which results in failing to address the data sparsity in larger spaces or \textit{over-exploiting in small spaces}. In contrast, regression oracles like SquareCB adapts their exploration strategies to the action space, focusing more on larger action spaces where data scarcity is more pronounced. This behavior is visualized in Figure~\ref{fig:exploration-action-size}, which shows how SquareCB allocates exploration more effectively than $\epsilon$-greedy across different action space sizes.
Uniform random exploration in the small-action-space regime is the cause of a phenomenon we're calling \textit{effective exploration}. The \textit{effective exploration} rate—defined as the percentage of instances where the action chosen was not the optimal action—was often lower than the nominal exploration rate (\( \epsilon \)) in $\epsilon$-greedy policies. This discrepancy arises because there is a \( 1/N_A \) chance of selecting the best action randomly, where \( N_A \) is the size of the action space. For example, if $N_A=2$ then effective exploration of a 1\% $\epsilon$-greedy policy could be much lower than the expected 1\% due to the high probability of selecting the optimal action during exploration.

\begin{table}[h!]
    \centering
    \begin{tabular}{|c|c|} \hline 
         \textbf{Variant} & \textbf{Effective Exploration \%} \\
         \hline 
         SquareCB LR1k& 13.0\%\\ \hline 
         SquareCB LR 4K& 6.5\% \\ \hline 
         SquareCB LR 7K& 4.6\% \\ \hline 
         SquareCB LR 10K& 3.6\% \\ \hline 
         SquareCB LR 50K& 1.2\% \\ \hline 
         $\epsilon$-Greedy (6\%) & 3.41\% \\ \hline 
         $\epsilon$-Greedy (4\%) & 2.3\% \\ \hline 
         $\epsilon$-Greedy (1\%) & 0.7\% \\ \hline
    \end{tabular}
    \caption{Effective Exploration Rates for SquareCB and $\epsilon$-Greedy Policies. Counterintuitively, the $\epsilon$-greedy policies are exploring less than than expected due to the dynamics of the small action space regime. }
    \label{tab:effective-exploration}
\end{table}

Given the dynamic action space in our setup, understanding effective exploration rate was crucial for assessing its contribution to training data diversity. Table~\ref{tab:effective-exploration} summarizes the effective exploration rates for SquareCB and $\epsilon$-greedy variants.

\subsection{Action Diversity}

As we have verified the expected exploration improvements promised by regression oracles, now we'd like to look at a common problem with polices in general: popularity bias, or framed in another perspective: action diversity.

Maintaining action diversity is critical for training robust next-generation policies. In our setup, dynamic action spaces inherently introduce biases, as certain actions are only available in specific contexts. This context-based action restriction creates an imbalance in the action distribution.

\begin{figure}[h!]
    \centering
    \includegraphics[width=1.0\linewidth]{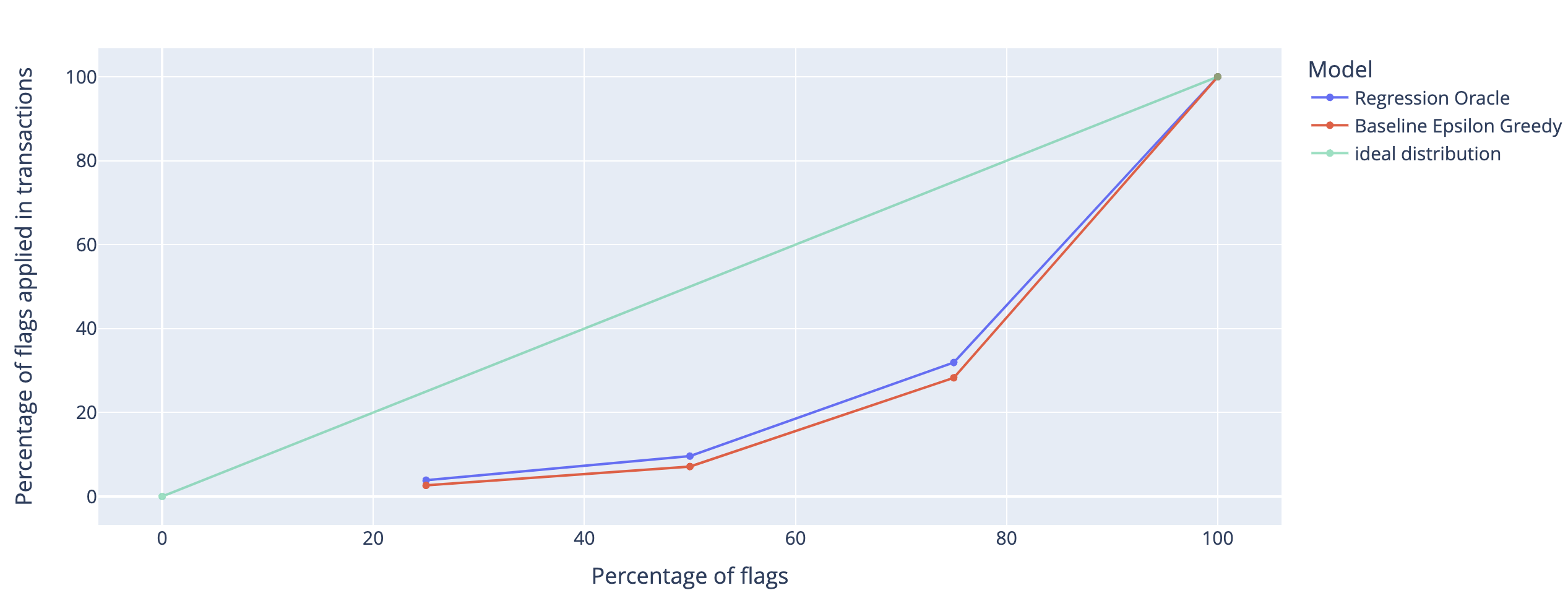}
    \caption{Lorenz Curve for contexts with 4 possible actions. (The word "flags" in the chart can be interpreted as "actions".)}
    \label{fig:lorenz-4-actions}
\end{figure}

\begin{figure}[h!]
    \centering
    \includegraphics[width=1\linewidth]{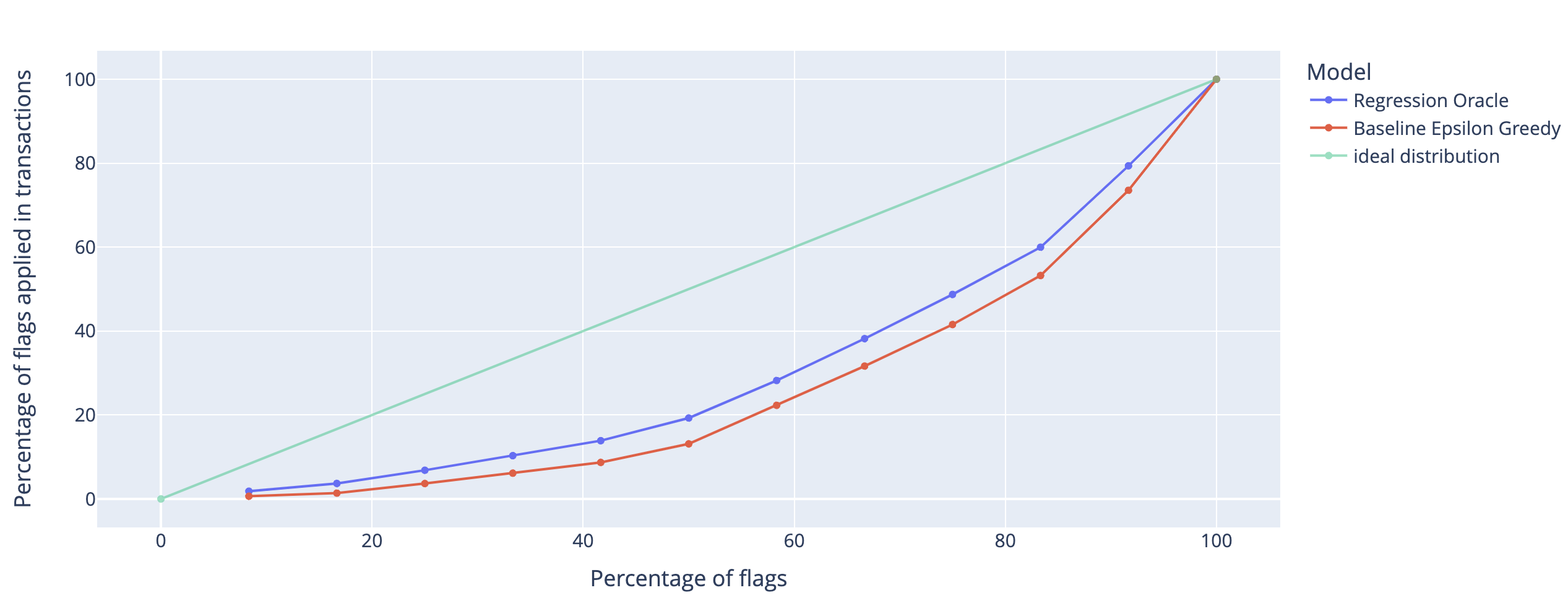}
    \caption{Lorenz Curve for contexts with 12 possible actions. (The word "flags" in the chart can be interpreted as "actions".)}
    \label{fig:lorenz-12-actions}
\end{figure}

To evaluate action diversity, we measured Lorenz curves and Gini coefficients across various context groups. Figures~\ref{fig:lorenz-4-actions} and \ref{fig:lorenz-12-actions} show improvements in action diversity for SquareCB compared to $\epsilon$-greedy policies. \textit{SquareCB effectively diversified action selection in larger action spaces while maintaining performance in simpler contexts.}

\begin{figure}[h!]
    \centering
    \includegraphics[width=1\linewidth]{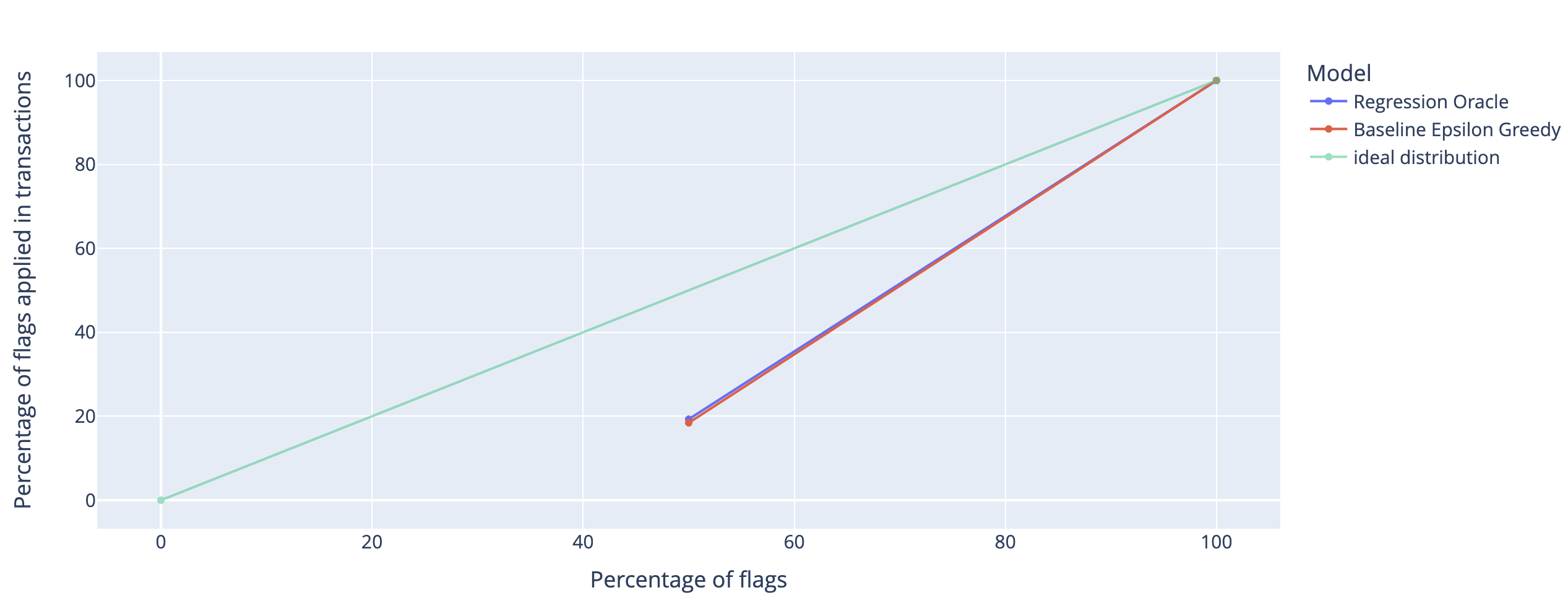}
    \caption{Lorenz Curve for contexts with 2 possible actions. (The word "flags" in the chart can be interpreted as "actions".)}
    \label{fig:lorenz-2-actions}
\end{figure}

Interestingly, in simpler contexts with only two possible actions, SquareCB and $\epsilon$-greedy policies showed nearly identical Lorenz curves (Figure~\ref{fig:lorenz-2-actions}). \textit{This indicates that SquareCB allocates exploration where it is needed most, leaving low-dimensional action spaces largely unaffected.}

Table~\ref{tab:gini-coefficients} provides the traffic-weighted Gini coefficients for each policy, showing that \textit{SquareCB reduced action inequality compared to the baseline $\epsilon$-greedy policy}.

These results indicate that SquareCB is better able to diversify action selection, especially in contexts with larger action sets. Nonetheless, the skewed exploration may lead to an imbalance in training data (fewer negative labels), which is discussed in Section~\ref{sec:eff-exp}.

\begin{table}[h!]
    \centering
    \begin{tabular}{|c|c|}
        \hline 
        \textbf{Model} & \textbf{Gini Coefficient} \\ \hline 
        \(\epsilon\)-Greedy & 0.39742 \\ \hline 
        SquareCB & 0.39265 \\ \hline
    \end{tabular}
    \caption{Traffic-weighted Gini coefficients for action diversity (lower values indicate better diversity).}
    \label{tab:gini-coefficients}
\end{table}

\subsection{Class Imbalance}

\begin{figure}[h!]
    \centering
    \includegraphics[width=1\linewidth]{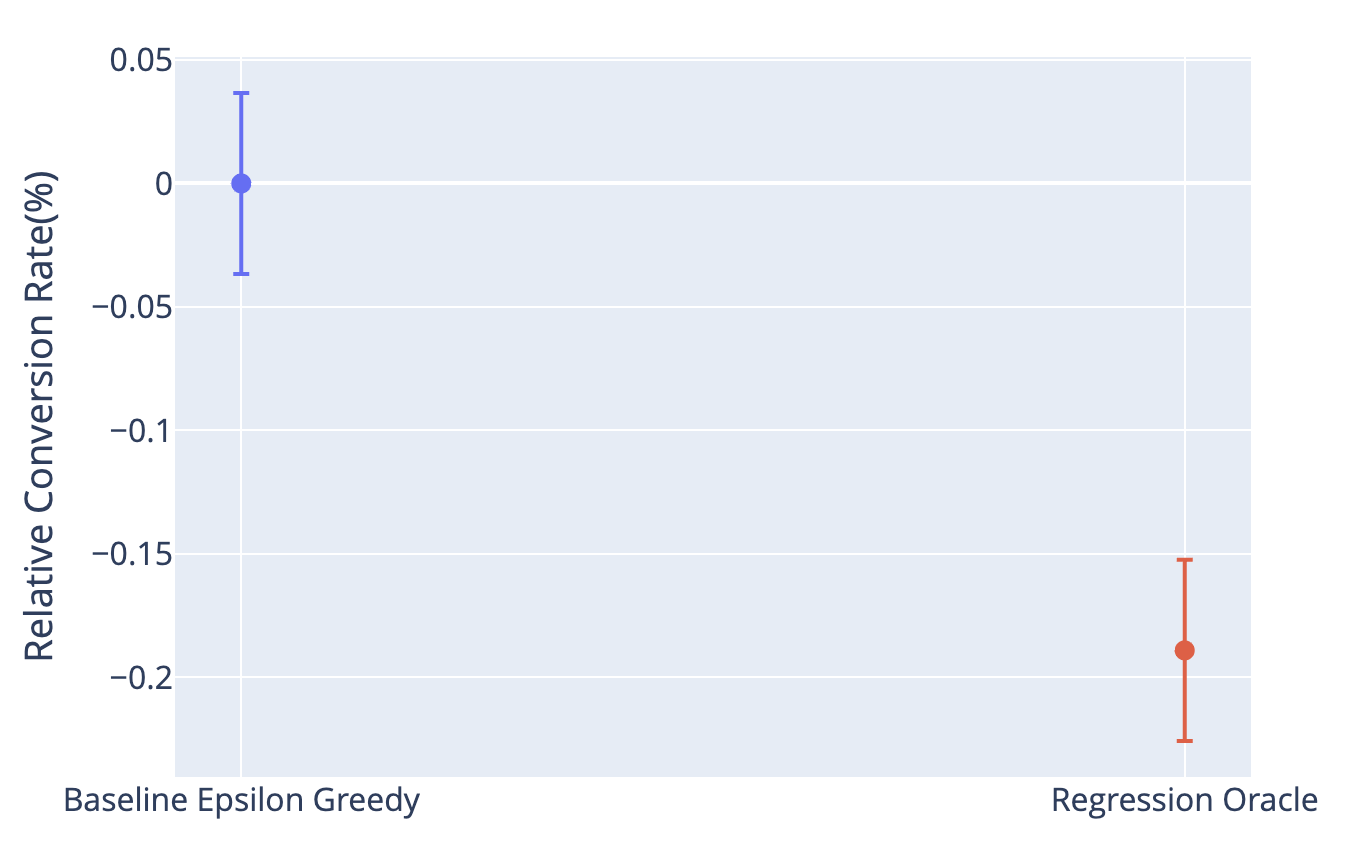}
    \caption{Comparison of two \(\epsilon\)-greedy policies trained on data from uniform and non-uniform (SquareCB) exploration. The performance drop in the SquareCB-based training data is attributed to an increased imbalance between positive and negative labels.}
    \label{fig:impact-training-data}
\end{figure}

A key finding of our study is that while the enhanced exploration of SquareCB improves immediate policy performance, it also exacerbates class imbalance in the logged data used for training future models (oracles). For instance, bandit feedback from the baseline \(\epsilon\)-greedy policies exhibits a class imbalance of approximately 87.4\% positive rewards. Under the SquareCB regime, this imbalance increases to 93.5\%, which correlates with a \textbf{0.2\%} performance regression (see Figure~\ref{fig:impact-training-data}).

As SquareCB improves the success rate of exploration actions, the frequency of negative outcomes (i.e., failed actions) decreases. These negative labels are essential for supervised models to learn robust decision boundaries between good and poor actions. Their reduction biases the training process, ultimately degrading the generalization capability of subsequent models. In other words, as the proportion of positive rewards increases, the resulting imbalance in the training data inadvertently compromises the quality of future models. This feedback loop highlights a fundamental challenge when applying ERM to logged bandit feedback. As current policies improve, the quality of training data for subsequent models deteriorates, resulting in weaker supervised models that eventually harm future policies. \textit{This paradox highlights a fundamental flaw in using ERM for bandit feedback.}

In addition, since the exploration is non-random, we can infer that the exploration data tend to be concentrated in the region of the higher action space size in this case of dynamic action spaces. Thus we lose the uniformity in data we have in epsilon greedy - of context and action combinations, which adds to the impact on future iterations of models trained on the logged feedback, explained further in the next section.

\subsubsection{Second Generation of Models}
\begin{figure}[h!]
    \centering
    \includegraphics[width=1\linewidth]{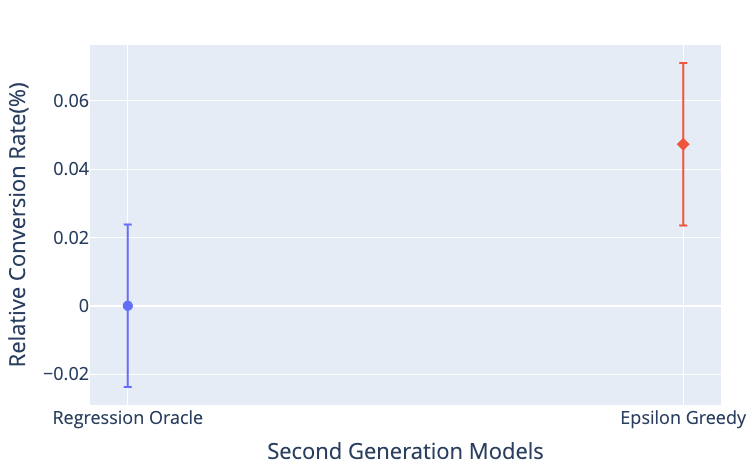}
    \caption{Performance of two second generation models resulting from Epsilon Greedy policy and Regression Oracle.  Despite Exploration sample being a significantly smaller part of the training data ($N_{exploitation}>>N_{exploration}$ ) 
    and exploitation being the majority in the training data of each, the Regression Oracle's second generation regresses in performance compared to Epsilon Greedy.}
    \label{fig:second_gen_models}
\end{figure}

As shown in Figure~\ref{fig:second_gen_models}, the second generation of models is
affected by the sampling skew introduced by the non-greedy selection rules of
SquareCB.  
The \emph{$\epsilon$-greedy} policy contributes a small share of uniformly sampled data,
enriching the next-generation training set with diverse triplets
\((\text{context}, \text{action}, \text{reward})\).
While SquareCB increases action diversity by occasionally selecting the
second- or third-best action, it still cannot match $\epsilon$-greedy’s uniformly
distributed \((\text{context}, \text{action})\) pairs and the resulting more
balanced reward distribution that benefits subsequent training iterations.

\section{Discussion}
\subsection{ERM for Bandit Feedback}

In our setting at Adyen, ERM leverages logged bandit feedback \((x, a, r)\), where \(r \in [0, 1]\), to learn the conditional distribution \(P(r \mid x, a)\). The goal is to predict the expected reward \(r\) for a given context-action pair \((x, a)\), enabling the policy to select actions that maximize expected rewards.

Empirical Risk Minimization (ERM) performs effectively when provided with ample and diverse training data. However, in the context of logged bandit feedback, only a limited subset of context-action pairs is observed, as \textit{counterfactual outcomes for unchosen actions remain unknown}. This scenario, commonly referred to as \textit{partial feedback}, introduces significant imbalances in the training data. Specifically, the logged bandit data tends to disproportionately represent actions with higher predicted probabilities of success, while actions with lower probabilities are underexplored. This imbalance distorts the learned conditional distribution $P(r \mid x, a)$, as the dataset lacks sufficient negative labels (representing low-reward actions). Consequently, the model struggles to effectively distinguish between high- and low-reward actions, impairing its ability to generalize to unseen or underexplored actions. This challenge arises because ERM inherently assumes a \textit{full-data setting}, an assumption that is fundamentally violated in the bandit feedback paradigm.

\section{Future Work}

\textbf{Counterfactual Risk Minimization (CRM)} provides an effective foundation for addressing the challenges of partial feedback in logged bandit data and represents the focus of our next line of research. By leveraging inverse propensity scoring (IPS) to reweight observed data, CRM addresses the inherent imbalances in logged feedback by assigning greater weight to underrepresented actions and mitigating the over-representation of high-reward actions, enabling more accurate reward estimation. Furthermore, CRM directly optimizes a counterfactual objective and employs variance reduction techniques, such as self-normalized estimators or clipping, to enhance stability and efficiency. Regularization methods further prevent overfitting to the reweighted data, ensuring better generalization to unseen or under-explored actions. By aligning the training objective with the counterfactual nature of logged bandit feedback, CRM offers a promising pathway to overcoming the limitations of ERM, paving the way for the development of robust, generalizable policies in settings with incomplete and biased feedback—an avenue we aim to explore in future work.

\section{Conclusion}
In this work, we addressed the challenge of leveraging logged bandit feedback to learn effective policies offline, a critical need in industry where untested policies cannot be fielded. While traditional contextual bandit methods rely on uniform random exploration or on-policy learning, we demonstrated the benefits of adopting regression oracles for non-uniform exploration, significantly reducing regret and improving performance.

\bibliographystyle{plain}
\bibliography{sample-base}

\begin{thebibliography}{10}

\bibitem{abe1999associative}
Naoki Abe and Philip~M. Long.
\newblock Associative reinforcement learning using linear probabilistic concepts.
\newblock In {\em Proceedings of the Sixteenth International Conference on Machine Learning}, pages 3--11. Morgan Kaufmann Publishers Inc., 1999.

\bibitem{grubhub}
Alex Egg.
\newblock Online learning for recommendations at grubhub.
\newblock In {\em Proceedings of the 15th ACM Conference on Recommender Systems}, RecSys '21, page 569–571, New York, NY, USA, 2021. Association for Computing Machinery.

\bibitem{foster18a}
Dylan Foster, Alekh Agarwal, Miroslav Dudik, Haipeng Luo, and Robert Schapire.
\newblock Practical contextual bandits with regression oracles.
\newblock In Jennifer Dy and Andreas Krause, editors, {\em Proceedings of the 35th International Conference on Machine Learning}, volume~80 of {\em Proceedings of Machine Learning Research}, pages 1539--1548. PMLR, 10--15 Jul 2018.

\bibitem{SquareCB}
Dylan Foster and Alexander Rakhlin.
\newblock Beyond {UCB}: Optimal and efficient contextual bandits with regression oracles.
\newblock In Hal~Daumé III and Aarti Singh, editors, {\em Proceedings of the 37th International Conference on Machine Learning}, volume 119 of {\em Proceedings of Machine Learning Research}, pages 3199--3210. PMLR, 13--18 Jul 2020.

\bibitem{SquareCB-lecture}
Akshay Krishnamurthy.
\newblock Lecture 4: Contextual bandits, cs, umass amherst, February 2022.

\bibitem{LinUCB}
Lihong Li, Wei Chu, John Langford, and Robert~E. Schapire.
\newblock A contextual-bandit approach to personalized news article recommendation.
\newblock In {\em Proceedings of the 19th International Conference on World Wide Web}, WWW '10, page 661–670, New York, NY, USA, 2010. Association for Computing Machinery.

\bibitem{adyen-badits}
Rodel van Rooijen.
\newblock Optimizing payment conversion rates with contextual multi-armed bandits, November 2020.

\bibitem{NNLinUCB}
Pan Xu, Zheng Wen, Handong Zhao, and Quanquan Gu.
\newblock Neural contextual bandits with deep representation and shallow exploration.
\newblock In {\em International Conference on Learning Representations}, 2022.

\bibitem{EE-Net}
Pan Xu, Zheng Wen, Handong Zhao, and Quanquan Gu.
\newblock Neural contextual bandits with deep representation and shallow exploration.
\newblock In {\em International Conference on Learning Representations}, 2022.

\bibitem{NeuralUCB}
Dongruo Zhou, Lihong Li, and Quanquan Gu.
\newblock Neural contextual bandits with {UCB}-based exploration.
\newblock In Hal~Daumé III and Aarti Singh, editors, {\em Proceedings of the 37th International Conference on Machine Learning}, volume 119 of {\em Proceedings of Machine Learning Research}, pages 11492--11502. PMLR, 13--18 Jul 2020.

\bibitem{Zhu2009RevenueOW}
Yunzhang Zhu, Gang Wang, Junli Yang, Dakan Wang, Jun Yan, and Zheng Chen.
\newblock Revenue optimization with relevance constraint in sponsored search.
\newblock In {\em KDD Workshop on Data Mining and Audience Intelligence for Advertising}, 2009.

\end{thebibliography}

\appendix
\newpage
\section{Appendix}

\subsection{Setting the Exploration Percentage}
Upon tuning the learning rate and finding the initial selection of variants we wanted to experiment with, we also found the corresponding exploration percentages we would gain if we used these learning rates for the SquareCB policy.
\begin{table}[h!]
    \centering
    \begin{tabular}{|c|c|} \hline 
         \textbf{Learning Rate}& \textbf{Exploration Percentage}\\ \hline 
         10K& 3.7\%\\ \hline 
         7K& 4.6\%\\ \hline 
        4K& 6.5\%\\ \hline 
        1K& 13\%\\ \hline
    \end{tabular}
    \caption{Exploration Percentage Guaranteed by various learning rates for SquareCB}
    \label{tab:exp-rates}
\end{table}

\subsection{Learning Rate Tuning}
The SquareCB policy, has a tunable parameter $\gamma$, the learning rate. Essentially this controls the amount of emphasis we wish to apply in the probability distribution calculation of squareCB:
\begin{equation}\label{1}
p_{t}(a) = \frac{1}{A + \gamma(\hat{y}_{t,b_{t}} - \hat{y}_{t,a_{t}})} 
\end{equation}
Where $\hat{y}_{t,b_{t}}$ is the predicted probability for the best action and $\hat{y}_{t,a_{t}}$ is the predicted probability of every other action. Therefore, $\gamma$ controls the importance given to the distance of an action from the best action \cite{SquareCB}. Therefore we tune the probability of selecting an action keeping in mind:
the lower the learning rate, the closer the probability of selecting a certain action moves to uniform selection, the higher the learning rate, the more this probability depends on the distance from the predicted probability of success of the best action.

An initial selection of models for experimentation was done through tuning of the learning rate - by observing the distribution of the "predicted probability of success" of the action selected, by the supervised classification model with a range of learning rates. Given the large volume of transactions at Adyen, the goal was to ensure that the predicted probability of success distribution of the SquareCB Policy was as close to a 100\% greedy policy.  For reference, this is how the distribution of probability of success of selection action looked for a learning rate of 1000:

\begin{figure}[h!]
    \centering
    \includegraphics[width=1.0\linewidth]{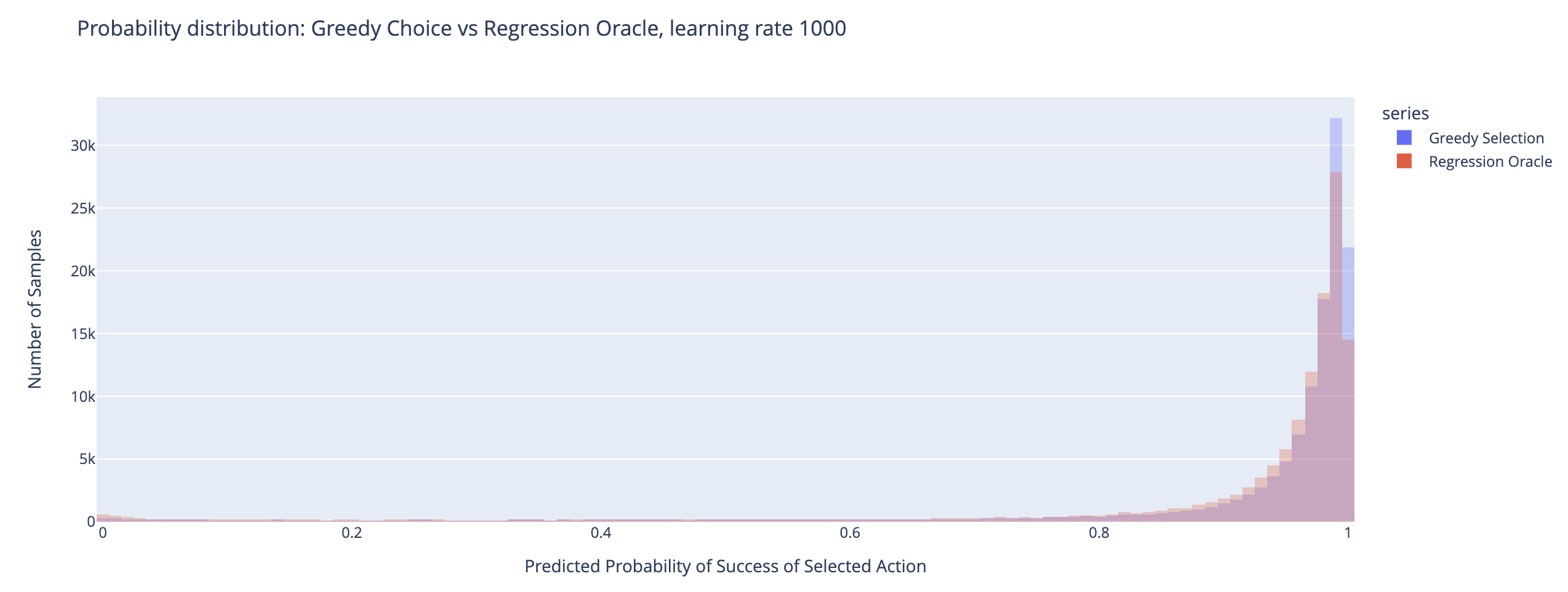}
    \caption{Probability distribution of selection actions $a$ for a purely greedy/exploitation policy vs a square-cb policy. }
    \label{fig:lr1k-dist}
\end{figure}

And this is how it looked for a learning rate of 10000:

\begin{figure}[h!]
    \centering
    \includegraphics[width=1.0\linewidth]{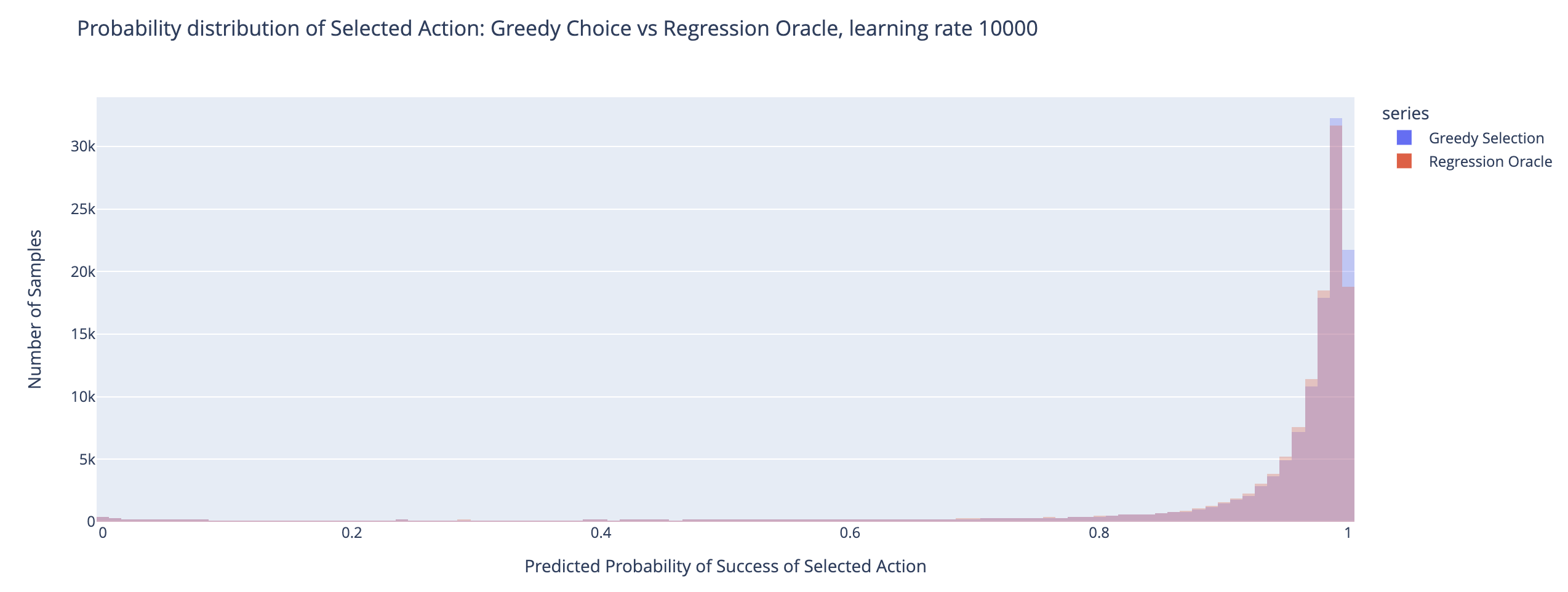}
    \caption{Selected Action Probability Distribution with learning rate 10000}
    \label{fig:enter-label}
\end{figure}
The selection of learning rate was done based on this probability distribution, exploration percentage and performance in A/B testing over a period of experimentation.

\subsection{Hyperparams}
\subsubsection{Model Configuration}
We implemented our regression oracle using XGBoost with the following key hyperparameters:

\begin{table}[h]
    \centering
    \caption{XGBoost Hyperparameters}
    \label{tab:hyperparams}
    \begin{tabular}{lll}
        \toprule
        \textbf{Parameter} & \textbf{Value} & \textbf{Description} \\
        \midrule
        learning\_rate & 0.1 & Step size shrinkage\\
        max\_depth & 10& Maximum tree depth \\
        subsample & 0.8 & Fraction of samples per tree \\
        colsample\_bytree & 0.8 & Fraction of features per tree \\
        n\_estimators & 1000 & Number of boosting rounds \\
        objective & binary:logistic & Loss function\\
        \bottomrule
    \end{tabular}
\end{table}

Training required approximately \textbf{4 hours per model} on 32 CPU cores, with periodic retraining every 7 days to maintain freshness.

\subsubsection{Feature Engineering}
Our feature pipeline included:
\begin{itemize}
    \item \textbf{Contextual features}: Transaction amount, currency, country, merchant category, device type
    \item \textbf{Temporal features}: Rolling 7-day authorization rates, time since last transaction
    \item \textbf{Embedding-based features}: Merchant representations learned via historical transaction patterns
    \item \textbf{Normalization}: Min-max scaling for monetary amounts
    \item \textbf{Imputation}: Median values for missing numeric features, special category for missing categoricals
\end{itemize}

\end{document}